\documentclass[]{spie}  

\pdfoutput=1  
\usepackage{amsmath,amsfonts,amssymb}
\usepackage{graphicx}
\usepackage[colorlinks=true, allcolors=blue]{hyperref}
\usepackage{multirow}
\usepackage{array}

\usepackage[table,xcdraw]{xcolor}

\begin{document} 
\title{An End-to-end Deep Learning Approach for Landmark Detection and Matching in Medical Images}

\author[a]{Monika Grewal}
\author[a]{Timo M. Deist}
\author[b]{Jan Wiersma}
\author[a, c]{Peter A. N. Bosman}
\author[b\thanks{\enspace Tanja Alderliesten is currently working at the Department of Radiation Oncology, Leiden University Medical Center,
P.O. Box 9600, 2300 RC Leiden, The Netherlands}]{Tanja Alderliesten}
\affil[a]{Life Sciences \& Health Research Group, Centrum Wiskunde \& Informatica, P.O. Box 94079, 1090 GB Amsterdam, The Netherlands}
\affil[b]{Department of Radiation Oncology, Amsterdam UMC, University of Amsterdam, P.O. Box 22660, 1100 DD  Amsterdam, The Netherlands}
\affil[c]{Faculty of Electrical Engineering, Mathematics and Computer Science, Delft University of Technology, P.O. Box 5, 2600 AA Delft, The Netherlands}

\authorinfo{Further author information: (Send correspondence to M.G.)\\
M.G.: Email: monika.grewal@cwi.nl}
\maketitle

\begin{abstract}
Anatomical landmark correspondences in medical images can provide additional guidance information for the alignment of two images, which, in turn, is crucial for many medical applications. However, manual landmark annotation is labor-intensive. Therefore, we propose an end-to-end deep learning approach to automatically detect landmark correspondences in pairs of two-dimensional (2D) images. Our approach consists of a Siamese neural network, which is trained to identify salient locations in images as landmarks and predict matching probabilities for landmark pairs from two different images. We trained our approach on 2D transverse slices from 168 lower abdominal Computed Tomography (CT) scans. We tested the approach on 22,206 pairs of 2D slices with varying levels of intensity, affine, and elastic transformations. The proposed approach finds an average of 639, 466, and 370 landmark matches per image pair for intensity, affine, and elastic transformations, respectively, with spatial matching errors of at most 1 mm. Further, more than 99\% of the landmark pairs are within a spatial matching error of 2 mm, 4 mm, and 8 mm for image pairs with intensity, affine, and elastic transformations, respectively. To investigate the utility of our developed approach in a clinical setting, we also tested our approach on pairs of transverse slices selected from follow-up CT scans of three patients. Visual inspection of the results revealed landmark matches in both bony anatomical regions as well as in soft tissues lacking prominent intensity gradients. 
\end{abstract}


\section{INTRODUCTION}
\label{sec:intro}  

Deformable Image Registration (DIR) can be extremely valuable in work-flows related to image-guided diagnostics and treatment planning. However, DIR in medical imaging can be challenging due to large anatomical variations between images. This is particularly the case in the lower abdomen, where internal structures can undergo large deformations between two scans of a patient due to physical conditions like presence of gas pockets and bladder filling. Such scenarios are particularly challenging for intensity based registration, as there are many local optima to overcome. Landmark correspondences between images can provide additional guidance information to the DIR methods\cite{alderliesten2015getting,HAN2015277} and increase the probabilty of finding the right transformation by adding landmark matches as an additional constraint or objective in the optimization. Since the manual annotation of anatomical landmarks is labor-intensive and requires expertise, developing methods for finding landmark correspondences automatically has great potential benefits.

The existing methods\cite{yang2017method,Werner2013,ForstnerOperator,Ghassabi2013,PIIFD} for obtaining landmark correspondences in medical images are based on large and time-consuming pipelines that involve identifying landmark locations followed by matching local feature descriptors\cite{Guo2016} within a restricted neighborhood. These methods rely upon multiple pre- and post-processing steps, multi-resolution search, and manual checking to achieve robustness; each step adding more heuristics and empirical hyperparameters to an already complex pipeline. Further, existing methods for landmark detection that restrict the definition of landmarks to certain intensity gradient patterns specific to the underlying data set or anatomical region may not be easily adaptable to other contexts \cite{hervella2018multimodal}. Generalizing the definition of landmarks and reducing the number of heuristics would allow for faster adaptation of automated methods for different clinical settings. In addition, faster execution times for landmark detection and matching could benefit their clinical application.

Recently, deep Convolutional Neural Networks (CNNs) have shown promising results for classification and segmentation tasks in medical imaging due to their capability of learning discriminant feature descriptors from raw images \cite{UnetMiccai, 10.1001/jama.2016.17216, esteva2017dermatologist}. There exist a few deep learning approaches for finding landmarks in medical images \cite{bier2018x, tuysuzoglu2018deep}. However, in these approaches a neural network is trained in a supervised manner to learn a small number of manually annotated landmarks. It is to be noted that a high density of landmark correspondences is desirable to effectively provide additional guidance to the DIR methods. In a supervised setting, it means annotating thousands of landmarks per CT scan, which is intractable in terms of required manual efforts. On the other hand, many deep learning approaches have been developed for automatically finding object landmarks in natural images \cite{Thewlis2017ICCV, zhang2018unsupervised, georgakis2018end, detone2018superpoint} that do not require manual annotations. Some of these approaches focus on discovering a limited number of landmarks in an image dataset. Whereas, others either fine-tune a pre-trained network or make use of incremental training in a self-supervised fashion.

Our proposed approach is based on the above-mentioned approaches developed for natural images and tailored to meet the specific requirements relating to the medical images. We propose a two-headed Siamese neural network that based on a pair of images simultaneously predicts the landmarks and their feature descriptors corresponding to each image. These are then sent to another module to predict their matching probabilities. We train the neural network from scratch and gradients are back-propagated from end-to-end. To the best of our knowledge, this is first endeavour to develop an end-to-end deep learning approach for finding landmark correspondences in medical images. Our approach has the following distinct advantages compared to existing methods for finding landmark correspondences:
\vspace{-3mm}
\begin{itemize}
\itemsep-0.2em
    \item[\textperiodcentered] Our approach is end-to-end deep learning based; therefore, the need for data pre- and post-processing during inference is avoided. In addition, the proposed approach is faster at run-time and has fewer hyperparameters than traditional approaches.
    \item[\textperiodcentered] We do not impose any prior on the definition of a landmark in an image. Instead, we train the network in a way that the landmarks represent salient regions in the image that can be found repeatedly despite potential intensity variations, and deformations.
    \item[\textperiodcentered] The proposed approach does not require manual annotations for training and learns from data in a self-supervised manner. 
    \item[\textperiodcentered] Our approach improves over the existing approaches for natural images by avoiding the need for pre-training, or incremental fine-tuning of the neural network.
\end{itemize}
\vspace{-3mm}

\section{MATERIALS AND METHODS}
\subsection{Data}
\label{sec:methods}

In total 222 lower abdominal Computed Tomography (CT) scans of female patients acquired for radiation treatment planning purposes were retrospectively included: 168 scans (24,923 two-dimensional (2D) slices) were used for training and 54 scans (7,402 2D slices) were used for testing. For a separate set of three patients, one original scan along with a follow-up CT scan was included. The scans of these three patients were used for testing the approach in a clinical setting. All CT scans had an in-plane resolution from 0.91 mm $\times$ 0.91 mm to 1.31 mm $\times$ 1.31 mm. All the 2D slices were resampled to 1 mm $\times$ 1 mm in-plane resolution.

\subsection{Approach}

\begin{figure} [htbp]
\begin{center}
\includegraphics[height=8cm, width=13cm]{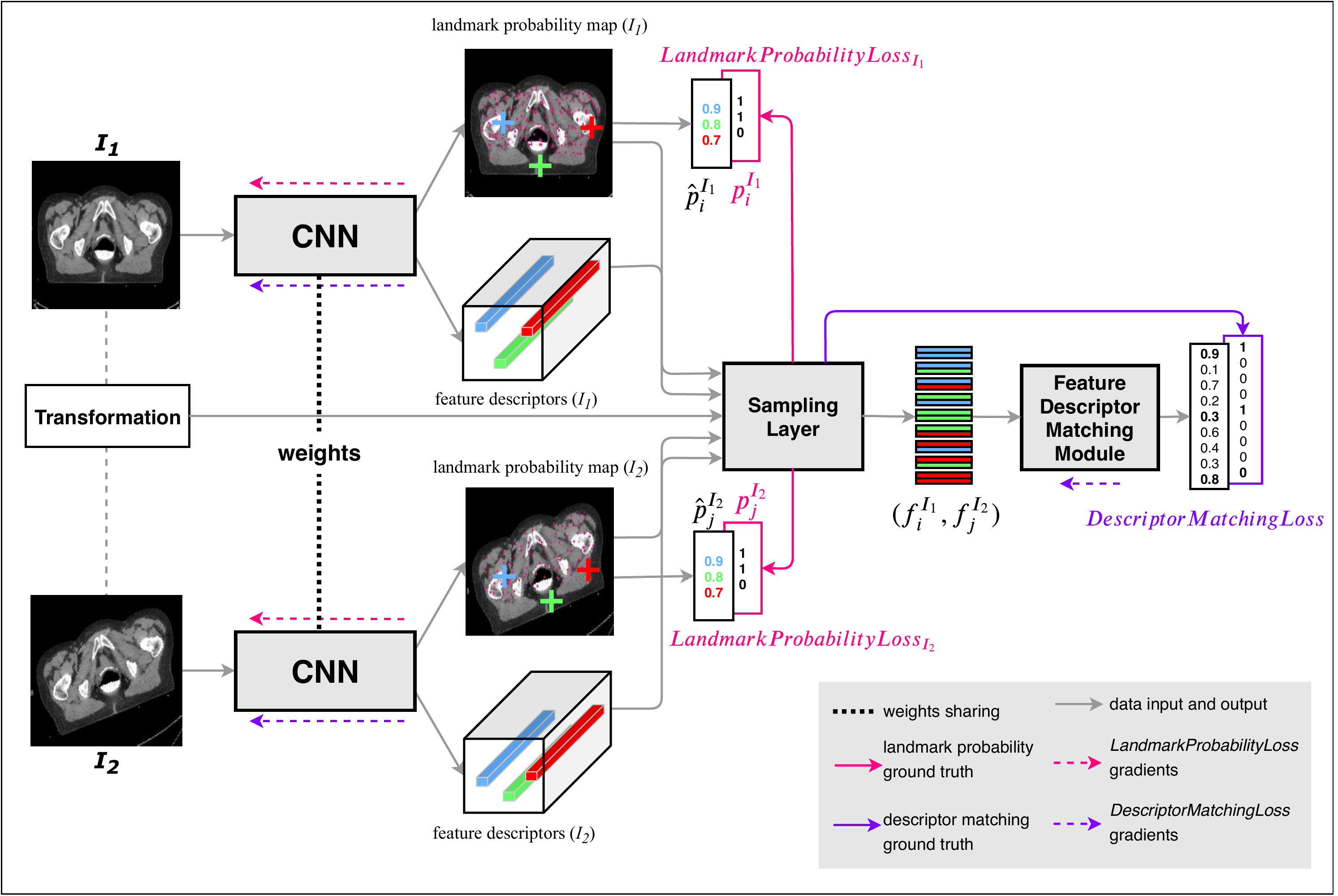}
\end{center}
\caption[example] 
{ \label{fig:architecture} 
\textbf{Schematic representation of our approach.}  
The weights are shared between two branches of the Siamese neural network. The transformation is required only during training for calculating the ground truths. Abbreviations of the data input and output at various stages follow the description in the text.}
\end{figure} 

In Figure \ref{fig:architecture}, the different modules of our approach are illustrated along with the data flow between them. Our approach comprises a Siamese architecture consisting of \textit{CNN branches} with shared weights. The outputs of the CNN branches are sent to a module named \textit{Sampling Layer} followed by another module named  \textit{Feature Descriptor Matching Module}. The network takes two images $I_1$ and $I_2$ as inputs and predicts $K_1$ and $K_2$ landmarks in $I_1$ and $I_2$, respectively. In addition, the network predicts matching probabilities ($\hat{c}_{i, j}$) for each landmark $i \in \{1, 2, ..., K_1\}$ in $I_1$ to a landmark $j \in \{1, 2, ..., K_2\}$ in $I_2$. In the following paragraphs, a description of each module is provided.

\subsubsection{CNN branches}
The CNN branches of the Siamese neural network have shared weights and consist of an encoder-decoder type network similar to the U-Net\cite{UnetMiccai} architecture. The only difference from the original implementation is that the number of convolutional filters in each layer is reduced by a factor of four to avoid overfitting. The implemented architecture contains 16, 32, 64, 128, and 256 convolutional filters in successive downsampling blocks respectively. The CNN branches give two outputs for each input image: a landmark probability map, and feature descriptors. The landmark probability map is computed at the end of the upsampling path after applying the sigmoid non-linearity and the feature descriptors are computed by concatenation of feature maps from the last two downsampling blocks. The feature maps from different downsampling blocks intrinsically allow for feature matching at multiple resolutions and abstraction levels.

\subsubsection{Sampling Layer}
The sampling layer is a parameter-free module of the network. It performs the following tasks:
\vspace{-3mm}
\begin{enumerate}
\itemsep-0.2em
    \item It samples $K_1$ and $K_2$ landmark locations in $I_1$ and $I_2$, respectively, which correspond to the highest probability score locations in the predicted landmark probability maps.
    \item It extracts predicted landmark probabilities $\hat{p}^{I_1}_i$, and $\hat{p}^{I_2}_j$ corresponding to $K_1$ and $K_2$ locations in landmark probability maps of image $I_1$ and $I_2$.
    \item It extracts feature descriptors $f^{I_1}_i$ and $f^{I_2}_j$ corresponding to the sampled landmark locations in $I_1$ and $I_2$, respectively, and creates feature descriptor pairs $(f^{I_1}_i, f^{I_2}_j)$ for each $i \in \{1, 2, ..., K_1\}$ and $j \in \{1, 2, ..., K_2\}$.
    \item During training, it generates the ground truths for landmark probabilities and feature descriptor matching probabilities on-the-fly as mentioned in Georgakis et al \cite{georgakis2018end}. Briefly, the sampled landmark locations of $I_2$ are projected onto $I_1$ based on the known transformation between the images. A landmark location $i$ in $I_1$ is decided to be matching to a landmark location $j$ in $I_2$ if the Euclidean distance between $i$ and the projection of $j$ on image $I_1$ is less than a predefined pixel threshold ($thresh_{pixels}$). 
\end{enumerate}
\vspace{-3mm}

\subsubsection{Feature Descriptor Matching Module}
All the feature descriptor pairs $(f^{I_1}_i, f^{I_2}_j)$ are fed to the feature descriptor matching module. The feature descriptor matching module consists of a single fully connected layer that predicts the matching probability for each feature descriptor pair.

\subsection{Training}
Training image pairs were generated on-the-fly by sampling a reference image randomly and generating the target image by transforming the reference image with a known transformation (randomly simulated brightness or contrast jitter, rotation, scaling, shearing, or elastic transformation). During training, the ground truths for landmark probabilities and feature descriptor matching probabilities are generated in the sampling layer as described above. We trained the network by minimizing a multi-task loss defined as follows:
\begin{equation}\label{eq1}
    Loss = {LandmarkProbabilityLoss}_{I_1} + {LandmarkProbabilityLoss}_{I_2} + {DescriptorMatchingLoss}
\end{equation}
The $LandmarkProbabilityLoss_{I_n}$ for the probabilities of landmarks in image $I_n, n \in \{1, 2\}$ is defined as:
\begin{equation}\label{eq2}
        {LandmarkProbabilityLoss}_{I_n} = \frac{1}{K_n} \sum_{i=1}^{K_n}\left((1 - \hat{p}^{I_n}_i) + CrossEntropy(\hat{p}^{I_n}_i, p^{I_n}_i)\right)
\end{equation}
where $CrossEntropy$ is the cross entropy loss between predicted landmark probabilities $\hat{p}^{I_n}_i$ and ground truths $p^{I_n}_i$.  The term $(1 - \hat{p}^{I_n}_i)$ in \eqref{eq2} encourages high probability scores at all the sampled landmark locations, whereas the cross entropy loss term forces low probability scores at the landmark locations that do not have a correspondence in the other image. As a consequence, the network is forced to predict high landmark probabilities only at the salient locations that have correspondence in the other image as well.

Hinge loss is widely used for learning discriminant landmark descriptors between matching and non-matching landmark pairs. We observed that a positive margin for the matching pairs in the hinge loss encourages the network to focus on hard positive examples (i.e., non-trivial landmark matches). Therefore, we defined $DescriptorMatchingLoss$ (equation \ref{eq3}) as a linear combination of hinge loss with a positive margin $m_{pos}$ on the L2-norm of feature descriptor pairs and cross entropy loss on matching probabilities predicted by the feature descriptor matching module.
\begin{equation}\label{eq3}
\begin{split}
DescriptorMatchingLoss & = \sum_{i=1, j=1}^{K_1, K_2} \left( \frac{c_{i, j} max(0, ||f^{I_1}_i - f^{I_2}_j||^2 - m_{pos})}{K_{pos}} \right. \\
        & + \frac{(1 - c_{i, j}) max(0, m_{neg} - ||f^{I_1}_i - f^{I_2}_j||^2)}{K_{neg}}\\
         & + \left. \frac{WeightedCrossEntropy(\hat{c}_{i, j}, c_{i, j})}{(K_{pos} + K_{neg})}\right)
\end{split}
\end{equation}
where $\hat{c}_{i, j}$, and $c_{i, j}$ are the predicted and the ground truth matching probabilities, respectively, for the feature descriptor pair $(f^{I_1}_i, f^{I_2}_j)$; $K_{pos}$ and $K_{neg}$ are the number of matching (positive class) and non-matching (negative class) feature descriptor pairs; $m_{pos}$ and $m_{neg}$ are the margins for the L2-norm of matching and non-matching feature descriptor pairs. $WeightedCrossEntropy$ is the binary cross entropy loss where the loss corresponding to positive class is weighted by the frequency of negative examples and vice versa. The gradients are back-propagated from end-to-end as indicated by the dashed arrows in Figure \ref{fig:architecture}.

\subsection{Constraining Landmark Locations}
A naive implementation of the approach may find all the landmarks clustered in a single anatomical region, which is not desirable. Therefore, to learn landmarks in all anatomical regions during training, we sample the landmarks on a coarse grid in the sampling layer, i.e., in each $8\times8$ pixel section of the grid, only one landmark location with the maximum landmark probability is sampled.

Another challenge in the CT scan imaging data comes from a large number of pixels belonging to the background. Traditionally, the image is cropped to the center to avoid prediction of landmarks in the background or on the patient table. However, this strategy requires an additional pre-processing step during inference. To avoid this, we computed a \textit{valid mask} for each image, which contained the value 1 at the location of body pixels and 0 elsewhere. The valid mask was generated by image binarization using intensity thresholding and removing small connected components in the binarized image. The network is trained to predict high landmark probabilities as well as feature descriptor matching probabilities only in the matching locations that correspond to a value of 1 in the valid mask. This allows the network to learn a content-based prior on the landmark locations and avoids the need for image pre-processing during inference.

\subsection{Inference}
During inference, only the locations in $I_{1}$ and $I_{2}$ with landmark probabilities above a threshold ($thresh_{landmark}$) are considered. Further, landmark pairs from different images are only matched if their matching is inverse consistent.  Suppose, locations $i\in\{1,..,K_{1}\}$ in $I_1$ and locations $j\in\{1,..,K_{2}\}$  in $I_2$ have landmark probabilities above $thresh_{landmark}$.
A pair $(i^{\ast},j^{\ast})$ is considered matching if there is no other pair $(i^{\ast},j')$ where $j'\in\{1,..,K_{2}\}$ or $(i',j^{\ast})$ where $i'\in\{1,..,K_{1}\}$ with higher descriptor matching probabilities or lower L2-norms for their feature descriptor pairs  $(f_{i^{\ast}}^{I_{1}},f_{j'}^{I_{2}})$ or $(f_{i'}^{I_{1}},f_{j^{\ast}}^{I_{2}})$.

\subsection{Implementation Details}
We implemented our approach using PyTorch\cite{paszke2017automatic}. We trained the network for 50 epochs using the Adam\cite{Kingma2014AdamAM} optimizer with learning rate $10^{-3}$ and a weight decay of $10^{-4}$. The training was done  with a batchsize of 4 and took 28 GPU (NVIDIA GeForce RTX 2080 Ti) hours. To allow for batching, a constant $K$ (set to 400) landmarks were sampled from all the images. The threshold for Euclidean distance while generating the ground truth ($thresh_{pixels}$) was 2 pixels. The margin for the L2-norm of matching feature descriptors ($m_{pos}$) was set to 0.1 and the margin for the L2-norm of non-matching pairs ($m_{neg}$) was set to 1. During inference, $thresh_{landmark}$ = 0.5 was used.

The empirical values for the hyperparameters were decided based on experience in the preliminary experiments. For example, the number for landmarks to be sampled during training ($K$) was decided such that the entire image was covered with sufficient landmark density, which was inspected visually. Similarly, the decision for $thresh_{pixels}$ was motivated by the fact that a threshold less than 2 pixels did not yield any matching landmarks in the first few iterations of the training and hence the network could not be trained. We initially trained the network with default values of $m_{pos}$, and $m_{neg}$ ($m_{pos} = 0$, and $m_{neg} = 1$). However, we noticed on the validation set that all the predicted landmark pairs were clustered in regions of no deformation. To avoid this behaviour, we trained the network with $m_{pos} = 0.1$ and $m_{pos} = 0.2$ so that the gradients were not affected by the hinge loss corresponding to easy landmark matches. The final results are reported corresponding to the run with $m_{pos} = 0.1$ as it had a better trade off between number of landmarks per image pair and difficulty of landmark locations. The value of $thresh_{landmark}$ was chosen to give the best trade off between the number of landmarks per image pair and the spatial matching error on the validation set.

\section{Experiments}
\label{sec:experiments}
\subsection{Baseline}
Scale Invariant Feature Transform (SIFT\cite{lowe2004distinctive}) based keypoint detectors and feature descriptors are prevalent approaches used in both natural image analysis as well as in medical image analysis \cite{Ghassabi2013}. Therefore, we used the OpenCV\cite{opencvlibrary} implementation of SIFT as the baseline approach for comparison. We used two matching strategies for SIFT:
a) brute-force matching with inverse consistency (similar to our approach, we refer to this approach as SIFT-InverseConsistency), b) brute-force matching with ratio test (as described in the original paper\cite{lowe2004distinctive}, we refer to this approach as SIFT-RatioTest). Default values provided in the OpenCV implementation were used for all other hyperparameters.

\subsection{Datasets}
The performance is evaluated on two test sets. First, for quantitative evaluation, we transformed all 7,402 testing images from 54 CT scans with three different types of transformations corresponding to intensity (jitter in pixel intensities = $\pm20\%$ maximum intensity), affine (pixel displacement: median = 29 mm, Inter Quartile Range (IQR) = 14 mm - 51 mm), and elastic transformations (pixel displacement: median = 12 mm, IQR = 9 mm - 15 mm), respectively. Elastic transformations were generated by deforming the original image according to a deformation vector field representing randomly-generated 2D Gaussian deformations. The extent of transformations was decided such that the intensity variations and the displacement of pixels represented the typical variations in thoracic and abdominal CT scan images \cite{LiverDeformation, polzin2013combining}. This resulted in three sets of 7,402 2D image pairs (total 22,206 pairs).

Second, to test the generalizability of our approach in a clinical setting, image pairs were taken from two CT scans of the same patient but acquired on different days. The two scans were aligned with each other using affine registration in the SimpleITK \cite{lowekamp2013design} package. This process was repeated for three patients.

\subsection{Evaluation}
For quantitative evaluation, we projected the predicted landmarks in the target images to the reference images and calculated the Euclidean distance to their corresponding matches in the reference images. We report the cumulative distribution of landmark pairs with respect to the Euclidean distance between them.

The performance of our approach on clinical data was assessed visually. We show the predicted results on four transverse slices belonging to different anatomical regions. To visually trace the predicted correspondences of landmarks, the colors of the landmarks in both the images vary according to their location in the original CT slice. Similarly colored dots between slices from original and follow-up image represent matched landmarks.

\section{RESULTS}
\label{sec:results}
\begin{table}[]
\caption{\label{tab:table1} \textbf{Description of predicted landmark matches.} Median number of landmark matches per image pair with Inter Quartile Range (IQR) in parentheses are provided together with the spatial matching error. The entries in bold represent the best value among all approaches.}
\vspace{0.2cm}
\centering
\begin{tabular}{| m{2cm} | m{4cm} | m{2.8cm} m{2.8cm} m{2.8cm} |}
\hline
\rowcolor[HTML]{C0C0C0} 
\multicolumn{2}{|c|}{\textbf{Transformations}}              & \textbf{Intensity} & \textbf{Affine} & \textbf{Elastic} \\[2.0ex]\hline
\multirow{3}{2cm}{\textbf{No. of landmarks}}                      & Proposed Approach & 639 (547 - 729)                   & 466 (391 - 555)                & 370 (293 - 452) \\
                                                    & SIFT - InverseConsistency & \textbf{711 (594 - 862)}                   & \textbf{610 (509 - 749)}                & \textbf{542 (450 - 670)}                \\    
                                                    & SIFT - RatioTest & 698 (578 - 849)                   & 520 (426 - 663)                & 418 (330 - 541)  \\[0.2ex] \hline
\multirow{3}{2cm}{\textbf{Spatial matching error (mm)}} & Proposed Approach & \textbf{0.0 (0.0 - 0.0)}                   & \textbf{1.0 (0.0 - 1.4)}                & \textbf{1.0 (1.0 - 1.4)}  \\ 
                                                    & SIFT - InverseConsistency     & 1.0 (1.0 - 1.4)                   & 1.0 (1.0 - 1.4)                & 1.0 (1.0 - 2.0)    \\
                                                    & SIFT - RatioTest     & 1.0 (1.0 - 1.4)                   & 1.0 (1.0 - 1.4)                & \textbf{1.0 (1.0 - 1.4)} \\[0.2ex] \hline
\end{tabular}
\end{table}

The inference time of our approach per 2D image pair is within 10 seconds on a modern CPU without any parallelization. On the GPU the inference time is $\sim$20 milliseconds. The model predicted on average 639 (IQR = 547 - 729), 466 (IQR = 391 - 555), and 370 (IQR = 293 - 452) landmark matches per image pair for intensity, affine, and elastic transformations, respectively. 

\subsection{Simulated Transformations}
\begin{figure}[htbp]
\begin{center}
\includegraphics[height=5cm, width=17cm]{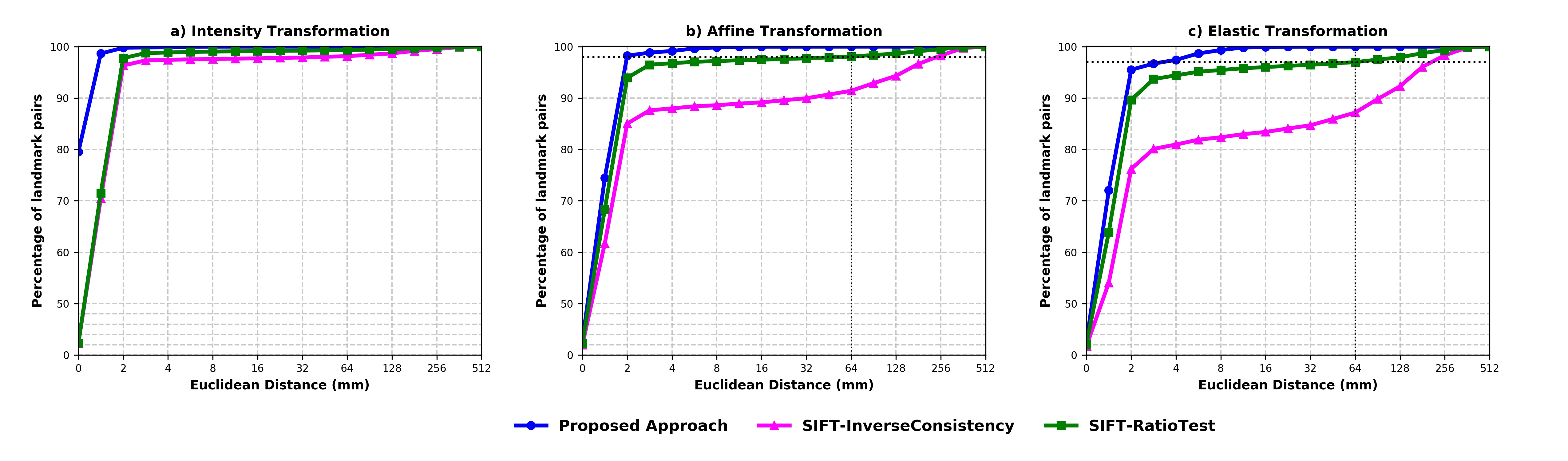}
\caption{\label{fig:pck} \textbf{Cumulative distribution of landmarks.} The cross-hairs in (b) and (c) correspond to the percentile of landmarks in SIFT-RatioTest at 64 mm.}
\end{center}
\end{figure}

Table \ref{tab:table1} describes the number of landmark matches per image pair and the spatial matching error for both our approach and the two variants of SIFT. Though our approach finds less landmarks per image as compared to the two variants of SIFT, the predicted landmarks have smaller spatial matching error than the SIFT variants. Further, Figure \ref{fig:pck} shows the cumulative distribution of landmark pairs with respect to the Euclidean distance between them. All the approaches are able to find more than 90\% of landmark matches within 2 mm error for intensity transformations. Predicting landmark correspondences under affine and elastic transformations is considerably more difficult; this can also be seen in the worse performance of all approaches. However, our approach is still able to find more than 99\% of landmark matches within a spatial matching error of 4 mm and 8 mm, respectively for affine and elastic transformations. However, a noticeable percentage (about 2\% for affine transformations and 3\% for elastic transformations) of landmarks detected by SIFT-RatioTest are wrongly matched with landmarks from far apart regions (more than 64 mm). It should be noted that if landmark matches with such high inaccuracies are used for providing guidance to a registration method, it may have a deteriorating effect on the registration if the optimizer is not sufficiently regularized.
For visual comparison, the landmark correspondences in pairs of original and elastic transformed images are shown in Figure \ref{fig:qualitative} (rows a-b) for our approach as well as for SIFT. As can be seen, the cases of mismatch in predictions from our approach (i.e., the number of landmarks in transformed slices not following the color gradient in the original slice) are rather scarce in comparison to the baseline approaches. Another interesting point to note is the difference in the landmark locations from our approach and the two baseline approaches. Since SIFT is designed to predict landmarks at locations of local extrema, the landmark matches are concentrated on the edges in the images. Our approach, however, predicts matches in soft tissue regions as well. Further inspection reveals that our approach predicts a considerable number of landmark matches even in the deformed regions in contrast to the baseline approaches. The capability to establish landmark correspondences in the soft tissues and deformed regions is important because DIR methods can especially benefit from guidance information in these regions.

\subsection{Clinical Transformations}
\begin{figure}[h]
\vspace{0.5cm}
\begin{center}
\includegraphics[height=13cm, width=16cm]{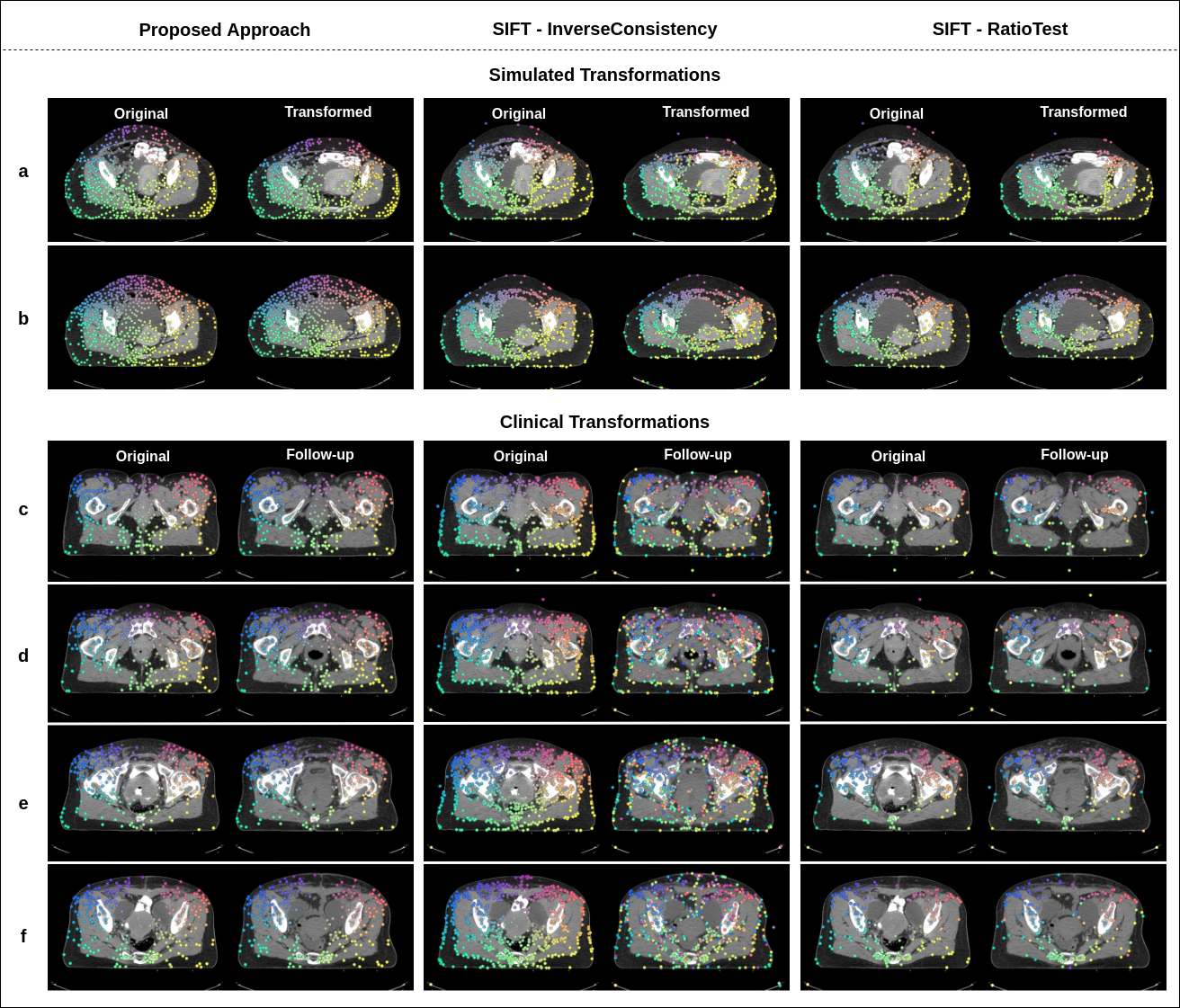}
\end{center}
\caption[example] 
{ \label{fig:qualitative} 
\textbf{Landmark correspondences for pairs of different transverse slices in abdominal CT scans.} 
 The landmark correspondences predicted by our approach are shown in comparison with two variants of SIFT. \textbf{Rows (a-b)} show predictions on pairs of original (left) and elastic transformed (right) slices. \textbf{Rows (c-f)} show transverse slices taken from different anatomical regions. The slices in the original CT (left) are matched with a similar slice from a follow-up CT scan (right) by affine registration.}
\end{figure}

Rows c-f in Figure \ref{fig:qualitative} show landmark correspondences in pairs of transverse slices corresponding to the lower abdominal region in the original and follow-up CT for our approach as well as for SIFT.  As can be seen, the original and follow-up slices have large differences in local appearance of structures owing to contrast agent, bladder filling, presence or absence of gas pockets, which was not part of the training procedure. It is notable that the model is able to find considerable landmark matches in image pairs despite these changes in local appearance. Moreover, the spatial matching error of landmarks seems similar to that of images with simulated transformations, in contrast to the baseline approach SIFT-InverseConsistency. Further, SIFT-RatioTest predicts fewer mismatched landmarks compared to SIFT-InverseConsistency, but this is achieved at the cost of a large decrease in the number of landmark matches per image pair. 

\section{DISCUSSION AND CONCLUSIONS}
\label{sec:conclusion}
With a motivation to provide additional guidance information for DIR methods of medical images, we developed an end-to-end deep learning approach for the detection and matching of landmarks in an image pair. To the best of our knowledge, this is the first approach that simultaneously learns landmark locations as well as the feature descriptors for establishing landmarks correspondences in medical imaging. While the final version of this manuscript was being prepared, we came across one research on retinal images \cite{truong2019glampoints}, whose approach for landmark detection using UNet architecture in a semi-supervised manner is partly similar to ours. However, our approach not only learns the landmark locations, but also the feature descriptors and the feature matching such that the entire pipeline for finding landmark correspondences can be replaced by a neural network. Therefore, our approach can be seen as an essential extension to the mentioned approach. 

Our proposed approach does not require any expert annotation or prior knowledge regarding the appearance of landmarks in the learning process. Instead, it learns landmarks based on their distinctiveness in feature space despite local transformations. Such a definition of landmarks is generic so as to be applicable in any type of image and sufficient for the underlying application of establishing correspondences between image pairs. Further, in contrast to the traditional unsupervised approaches for landmark detection in medical imaging, the proposed approach does not require any pre- or post-processing steps, and has fewer hyperparameters.   

The main challenge for intensity based DIR methods is to overcome local optima caused by multiple low contrast regions in the image, which result in image folding and unrealistic transformations in the registered image. It can be speculated that the availability of landmark correspondences in the low contrast image regions may prove to be beneficial for DIR methods. Moreover, a uniform coverage of entire image is desirable for improved performance. Upon visual inspection of the landmarks predicted by our approach, we observed that our approach not only finds landmark correspondences in bony anatomical regions but also in soft tissue regions lacking intensity gradients. Moreover, a considerable density of landmarks (approximately 400 landmarks per image pair) was observed despite the presence of intensity, affine, or elastic transformations. Based on these observations, we are optimistic about the potential added value of our approach to the DIR methods.

We validated our approach on images with simulated intensity, affine, and elastic transformations. The quantitative results show low spatial matching error of the landmarks predicted by our approach. Additionally, the results on clinical data demonstrate the generalization capability of our approach. We compared the performance of our approach with the two variants of widely used SIFT keypoint detection approach. Our approach not only outperforms the SIFT based approach in terms of matching error under simulated transformations, but also finds more accurate matches in the clinical data. As such the results look quite promising. However, the current approach is developed for 2D images i.e., it overlooks the possibility of the out-of-plane correspondences in two CT scans, which is quite likely especially in lower abdominal regions. The extension of the approach to 3D is, therefore, imperative so as to speculate into its benefits in providing additional guidance information to the DIR methods.

\section{ACKNOWLEDGEMENTS}
The research is part of the research programme, Open Technology Programme with project number 15586, which is financed by the Dutch Research Council (NWO), Elekta, and Xomnia. Further, the work is co-funded by the public-private partnership allowance for top consortia for knowledge and innovation (TKIs) from the Ministry of Economic Affairs.

\bibliographystyle{spiebib} 
\bibliography{main} 

\end{document}